# Surrogate compliance modeling enables reinforcement learned locomotion gaits for soft robots

Jue Wang[1†], Mingsong Jiang[1†], Luis A. Ramirez[1], Bilige Yang[1], Mujun Zhang[1], Esteban Figueroa[1], Wenzhong Yan[1], Rebecca Kramer-Bottiglio[1*]

[1]Department of Mechanical Engineering, Yale University, New Haven, CT 06511, USA.

*Corresponding author. Email: rebecca.kramer@yale.edu

†These authors contributed equally to this work.

**Adaptive morphogenetic robots adapt their morphology and control policies to meet changing tasks and environmental conditions. Many such systems leverage soft components, which enable shape morphing but also introduce simulation and control challenges. Soft-body simulators remain limited in accuracy and computational tractability, while rigid-body simulators cannot capture soft-material dynamics. Here, we present a surrogate compliance modeling approach: rather than explicitly modeling soft-body physics, we introduce indirect variables representing soft-material deformation within a rigid-body simulator. We validate this approach using our amphibious robotic turtle, a quadruped with soft morphing limbs designed for multi-environment locomotion. By capturing deformation effects as changes in effective limb length and limb center of mass, and by applying reinforcement learning with extensive randomization of these indirect variables, we achieve reliable policy learning entirely in a rigid-body simulation. The resulting gaits transfer directly to hardware, demonstrating high-fidelity sim-to-real**

**performance on hard, flat substrates and robust, though lower-fidelity, transfer on rheologically complex terrains. The learned closed-loop gaits exhibit unprecedented terrestrial maneuverability and achieve an order-of-magnitude reduction in cost of transport compared to open-loop baselines. Field experiments with the robot further demonstrate stable, multi-gait locomotion across diverse natural terrains, including gravel, grass, and mud.**

# 1  Introduction

Throughout evolution, animals have developed remarkable survival strategies, enabling them to thrive across diverse habitats. Species that inhabit a single ecological niche typically exhibit highly specialized body structures and gait kinematics, achieving efficient locomotion in their specific environments. Emerging shape-morphing materials and mechanisms present the opportunity to realize synthetic systems that can adapt between specialized morphologies and corresponding behavioral control policies, thereby enabling efficient task performance in multiple environments (*1, 2*). Drawing inspiration from biological morphogenesis, where an organism's shape develops and adapts over time in response to genetic programming and environmental cues, adaptive morphogenesis describes a robot's ability to change its physical form (morphology) in response to its environment or task demands (*3*).

Many emerging adaptive morphogenetic robots incorporate soft robotic materials and components, which can enable dramatic changes in morphology. For example, Kim *et al.* (*4*) demonstrated a locomotion robot with soft inflatable skin for adaptable buoyancy, volumetric shape, and physical compliance; Melancon *et al.* (*5*) presented inflatable and bistable origami structures for deployable morphologies; Kabutz *et al.* (*6*) developed a body-compliant multi-environment microrobot; and Baines *et al.* (*3*) introduced a quadrupedal robot with morphing limbs specialized for terrestrial and aquatic environments. While not strictly soft, Nygaard *et al.* (*7*) also demonstrated a quadruped with adaptable leg length for terrestrial multi-environment locomotion.

Progress has been made toward adaptive morphogenetic robots; however, the efficacy of morphological specialization is only realized through co-optimization of shape and gait. Most existing adaptive morphogenetic robots, and more broadly soft robots, remain pre-programmed with open-loop or hand-designed gaits (*8–12*). While these approaches demonstrate proof-of-concept locomotion, they seldom yield optimal gaits in terms of stability, agility, or energy efficiency.

Recent advances in machine learning—particularly reinforcement learning (RL)—offer a promising means to address shape–gait co-optimization. RL has transformed control of rigid-bodied legged robots, enabling agile and robust behaviors through scalable simulation-to-reality pipelines (*13–16*), with related work exploring Bayesian optimization for similar objectives (*17*). However, extending these methods to (partially) soft robots presents substantial challenges. Accurately simulating the interactions between soft bodies and their contact with substrates typically requires computationally intensive finite element models (*18*), which are impractical for the large-scale parallelization essential to RL. Conversely, simplified physics models often fail to capture the nuanced dynamics needed for high-performance gait learning (*19*). Although some RL pipelines incorporate terrain variability or physics-informed learning for rigid robots (*14, 20–27*), these methods are not directly applicable to soft or soft–rigid hybrid robots. As a result, most robots with soft components continue to rely on open-loop or scripted gaits, leading to suboptimal performance, especially on deformable or unstructured terrains (*28*).

Here, we present an approach to derive efficient locomotion gaits for robots containing soft components using RL. Rather than explicitly modeling soft-body physics, we introduce indirect variables representing soft-material deformation within a rigid-body simulator. This abstraction retains computational efficiency while enabling large-scale policy training. As a case study, we test this approach on our amphibious robotic turtle (ART) (*3, 29, 30*), a soft–rigid hybrid robot comprising a rigid central body with variable-stiffness morphing limbs. For its ability to learn near-optimal gaits and operate untethered, we refer to this version of the robot as the learning untethered amphibious robotic turtle (LUART).

LUART's soft limb deformation effects are parameterized by randomizing the effective limb lengths and centers of mass during virtual robot generation, thereby indirectly capturing geometric changes arising from limb compliance. Aggressive penalties for deviations in body height are incorporated into the RL reward function, establishing closed-loop height control and ensuring stable locomotion across diverse limb configurations. Additionally, domain randomization of motor parameters and environmental conditions enhance the trained policy's adaptability, enabling robust locomotion across varied terrains.

The learned policies substantially enhance LUART's locomotion stability, agility, and efficiency across a diverse repertoire of gaits, including omnidirectional locomotion, translational and rotational turning, and stable transitions between low-profile crawling and upright walking. We validate these capabilities through real-world experiments, achieving successful sim-to-real transfer of all learned gaits. Quantitative evaluation reveals that gait fidelity between simulation and reality depends on terrain complexity, with the highest correspondence observed on flat, hard ground. Remarkably, gaits trained entirely in simulation also function effectively on rheologically complex substrates such as mud, demonstrating the robustness of the learned control strategies. In sum, we show that indirectly modeling soft-material deformation effects within a rigid-body simulation framework enables reinforcement learning of versatile, transferable policies, validated through field deployments in natural environments.

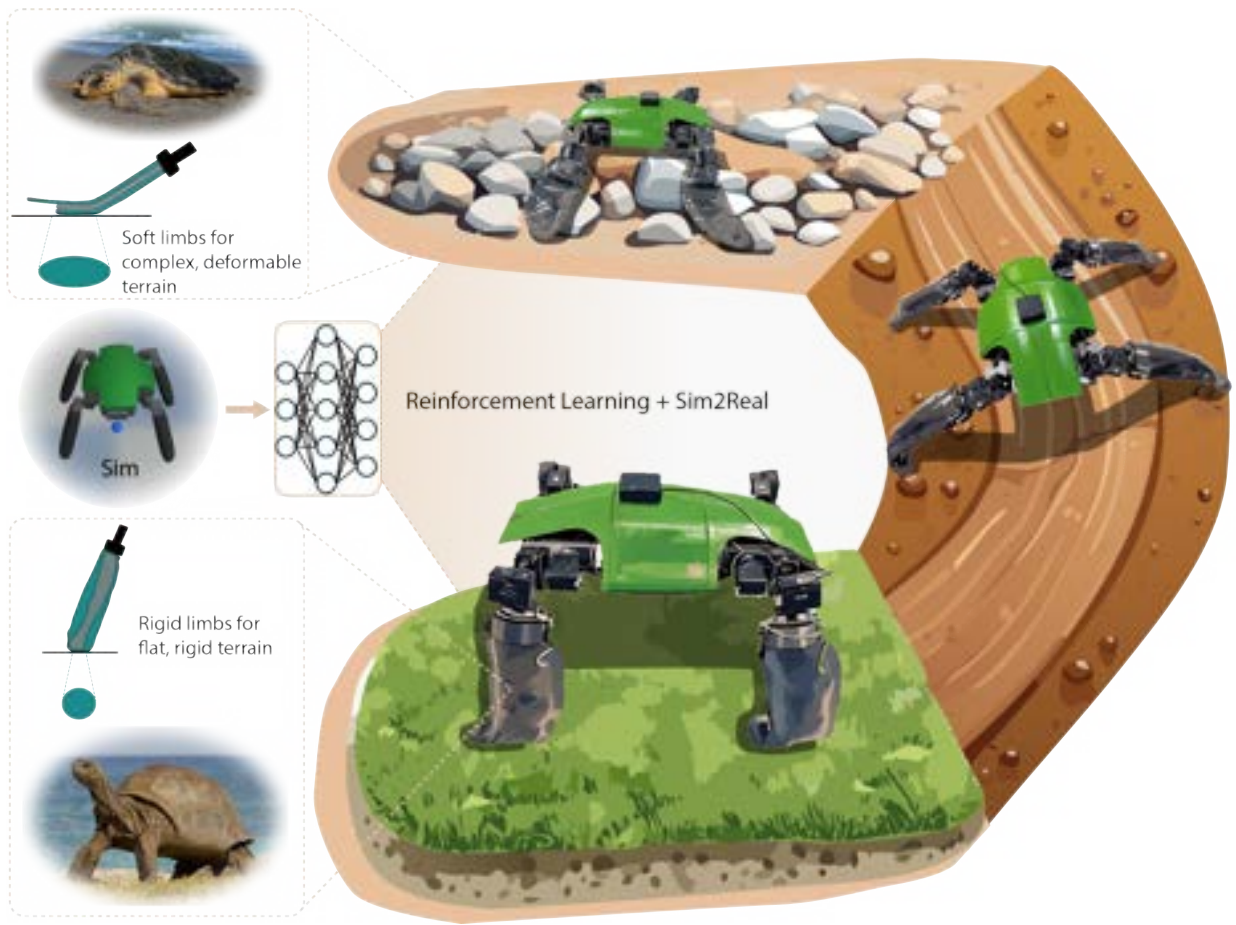

**Figure 1**: Surrogate compliance modeling enables reinforcement learning to optimize control policies for soft–rigid hybrid robots. Here, we use this approach to generate improved terrestrial gaits for a quadruped robot with variable-stiffness morphing limbs.

# 2 Results

## 2.1 Robot hardware

Our robot features a rigid main body and four morphing limbs, as illustrated in Fig. 1 and fig. S1. Each limb is connected to a shoulder joint composed of three motors, enabling independent control of pitch, roll, and yaw. The limbs comprise a central pneumatic pouch flanked by two rubber sheets embedded with laminar jamming layers. When inflated, the pouch expands into a cylindrical shape that stiffens the limb, mimicking the rigid, columnar legs of terrestrial tortoises. When deflated, the limb flattens into a hydrofoil-like profile, resembling the compliant flippers of sea turtles. The pouch provides the primary structural stiffness, while the jamming layers modulate buckling behavior under load.

The main body houses multiple onboard sensors, including a RealSense camera for visual and IMU data, a distance sensor for body-height feedback, a power sensor for energy monitoring, and a GPS module for localization. A compact pneumatic system modulates limb stiffness and enables dynamic shape change. All onboard computation and control are handled by a Jetson Orin Nano, and an onboard battery powers fully untethered operation.

We note that, unlike prior versions of our amphibious robot's limb design (*29, 30*), which relied on pneumatic inflation to drive shape change and exolayer jamming to lock or unlock those shapes, the morphing limb presented here uses fewer exolayers. As a result, the exolayer jamming has a negligible impact on the limb's compressive or bending stiffness when the pouch is inflated to any level (fig. S2 and section S1). In this design, limb stiffness is governed almost entirely by pouch pressurization and behaves in a largely binary manner: any amount of pressure produces a substantial increase in stiffness relative to the unpressurized state. When the pouch is uninflated (flipper mode) and unjammed, compression tests show that the buckling force is less than half the robot's body weight, resulting in gait failure (fig. S3). Jamming the exolayers (up to –11 psi), however, enhances stiffness sufficiently to enable successful upright walking. Likewise, in a cantilever test, maximum jamming approximately doubles the force required for buckling. In the robot, this corresponds to poor height control in the unjammed crawling condition and successful height control when jamming is applied (fig. S4; Movie S1).

## 2.2 Simulation to reality pipeline

Reinforcement learning (RL) is a powerful tool for discovering effective locomotion gaits, but it relies on high-fidelity robot simulations to ensure that policies trained in simulation transfer reliably to hardware. Because our robot uses soft robotic limbs, accurate modeling and simulation are particularly challenging. Existing soft-body simulators are computationally intensive, limiting scalability and making large-scale parallel RL training impractical. To address this, we introduce a method that captures the effects of material deformation by extensively randomizing variables that approximate limb compliance in a rigid-body simulator.

LUART features variable-stiffness, shape-morphing limbs. In a rigid-body simulator, limb compliance and geometry most directly influence three parameters: effective limb length (ELL), the limb center of mass ($CoM_l$), and body height from the ground. For example, in leg mode, both the overall limb stiffness (determined by pouch inflation and jamming pressure) and the limb angle relative to the body ($\theta$) influence the degree of limb bending, which in turn alters the ELL, $CoM_l$, and body height (Fig. 2A). In flipper mode, buckling yields comparable effects on these parameters.

Figs. 2B–C present tabulated measurements of ELL and corresponding body height across four limb states (inflated/deflated pouch; jammed/unjammed exolayers) and limb angles ($\theta$) ranging from 15° to 75°. Here, "inflated" denotes the maximum pouch inflation state, while "deflated" indicates complete deflation; "jammed" corresponds to the maximum jamming pressure, and "unjammed" refers to atmospheric pressure. The limbs can also achieve intermediate inflation and jamming levels.

These data show the broad range of ELLs exhibited by our robot and that the ELL of a limb is highly sensitive to its initial conditions, making it impractical to predict with precision. This observation motivates our approach: rather than assigning a fixed ELL, we randomize it during simulation to learn policies that are inherently robust to variations in limb deformation. Specifically, we vary the Unified Robot Description Format (URDF) limb length from 0.5 to 1.1 × the nominal value and shift the leg's center of mass by up to ±25% of its length about the midpoint, producing a diverse dataset of 12,000 robot models.

Training employs a Proximal Policy Optimization (PPO) algorithm within a student–teacher framework (asymmetric observations). The teacher policy is trained using privileged information to accelerate learning, while the student policy relies solely on onboard sensor data for real-world applicability; at deployment, only the student policy is used on hardware. The reward function promotes accurate tracking of commanded velocities and body height while penalizing instability, excessive energy consumption, and undesirable ground contacts. The use of softness surrogates and extensive domain randomization enables the learned policy to achieve relatively stable and agile locomotion in robots with soft and/or shape-changing components (Fig. 2D).

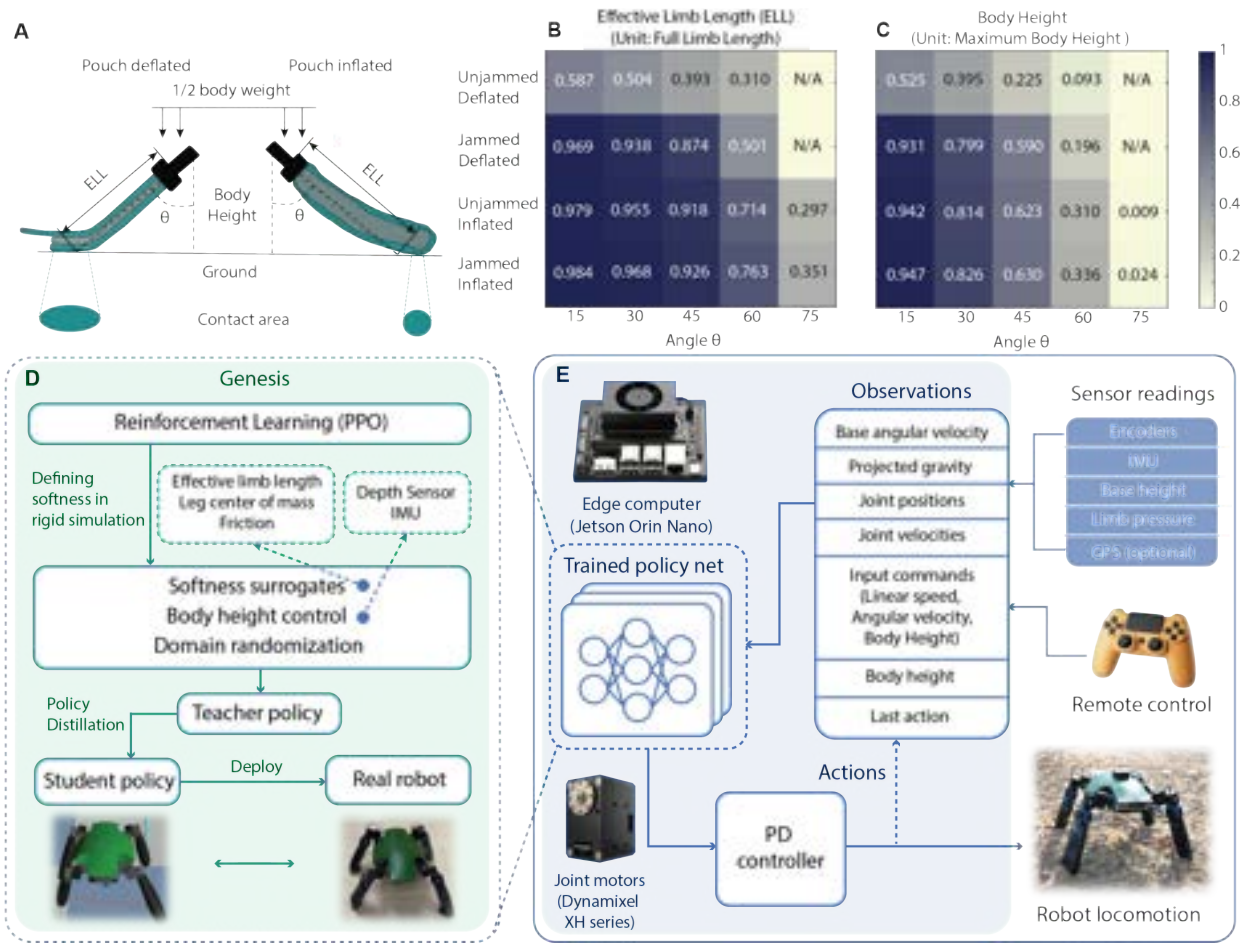

**Figure 2**: **Simulation-to-reality pipeline for LUART. A**, Two deformation modes of the soft limb. Deflation causes buckling and a pronounced reduction in effective limb length (ELL), while inflation induces slight bending (exaggerated here for clarity) that also alters ELL. **B**, ELL measured under four conditions (pouch de/inflated and exolayers un/jammed) at limb angles of $15°, 30°, 45°, 60°$, and $75°$. Values are expressed as fractions of the full limb length. **C**, Corresponding normalized body heights under the same conditions. **D**, Reinforcement learning (RL) pipeline in a rigid-body simulator (Genesis). Soft-limb effects were captured by randomizing surrogate parameters—effective limb length and limb center of mass—together with domain randomization and closed-loop body height control. A teacher policy was trained via Proximal Policy Optimization (PPO) using privileged information and distilled into a deployable student policy. **E**, Sim-to-real deployment. Sensor data (IMU, joint encoders, depth-based height) and remote commands were processed by an onboard policy network on a Jetson Orin Nano to generate motor targets executed through a PD controller. Limb pressures were regulated to control stiffness states.

For real-world deployment, sensor data are collected via ROS2 topics, including joint encoder positions and velocities, IMU-derived base angular velocity and projected gravity, and body height from depth sensors (Fig. 2E). These measurements, combined with remote control commands (linear speed, angular velocity, body height reference) and previous action states (motor target positions), form a comprehensive observation vector that is fed into the trained policy network. The network outputs updated motor target positions, which are executed through a PD controller driving Dynamixel XH-series motors. Simultaneously, limb pressures are monitored via pressure sensors to regulate stiffness states precisely, while real-time GPS and power sensors provide continuous monitoring of the robot's cost of transport (COT).

## 2.3 Simulation to reality evaluation

We evaluated sim-to-real performance of LUART's forward ($x$), lateral ($y$), and angular velocities for both walking and crawling gaits on flat, rigid ground (VCT flooring). Because this surface matches the terrain modeled in simulation, it provides a consistent baseline for comparing simulated and experimental behaviors. All tests were performed with fully inflated, jammed limbs (*i.e.*, "leg mode"), corresponding to the most rigid limb configuration and thus the closest match to the rigid-body simulation.

The proportional and derivative gains ($K_p$, $K_d$) of the PD controller and the action update frequency were identified as parameters for sim-to-real tuning. These were manually adjusted to achieve the closest correspondence between simulated and experimental locomotion on rigid ground. Robot speeds were computed from displacement-time trajectories and compared to their simulated counterparts (Fig. 3). Linear regression provided effective locomotion speeds, and the resulting slopes and relative errors are summarized in tables S1 (walking) and S2 (crawling).

For upright walking, optimal performance was achieved with $K_p = 550$, $K_d = 15$, and an action update frequency of 10 Hz. As shown in Fig. 3A, the robot was commanded to move forward and backward at velocities of 0.5, 0.75, and 1 (corresponding to 50%, 75%, and 100% of the simulated speed, respectively) in the $x$-direction, with positive and negative values indicating forward and backward motion. Dashed and solid lines denote simulated and experimental data, respectively, with shaded regions showing upper and lower bounds across five trials. Under these conditions, simulated and experimental steady-state speeds exhibit close quantitative agreement, with relative errors ranging from 2.29% to 7.76% (mean 5.29%) across all commanded forward and backward velocities (table S1).

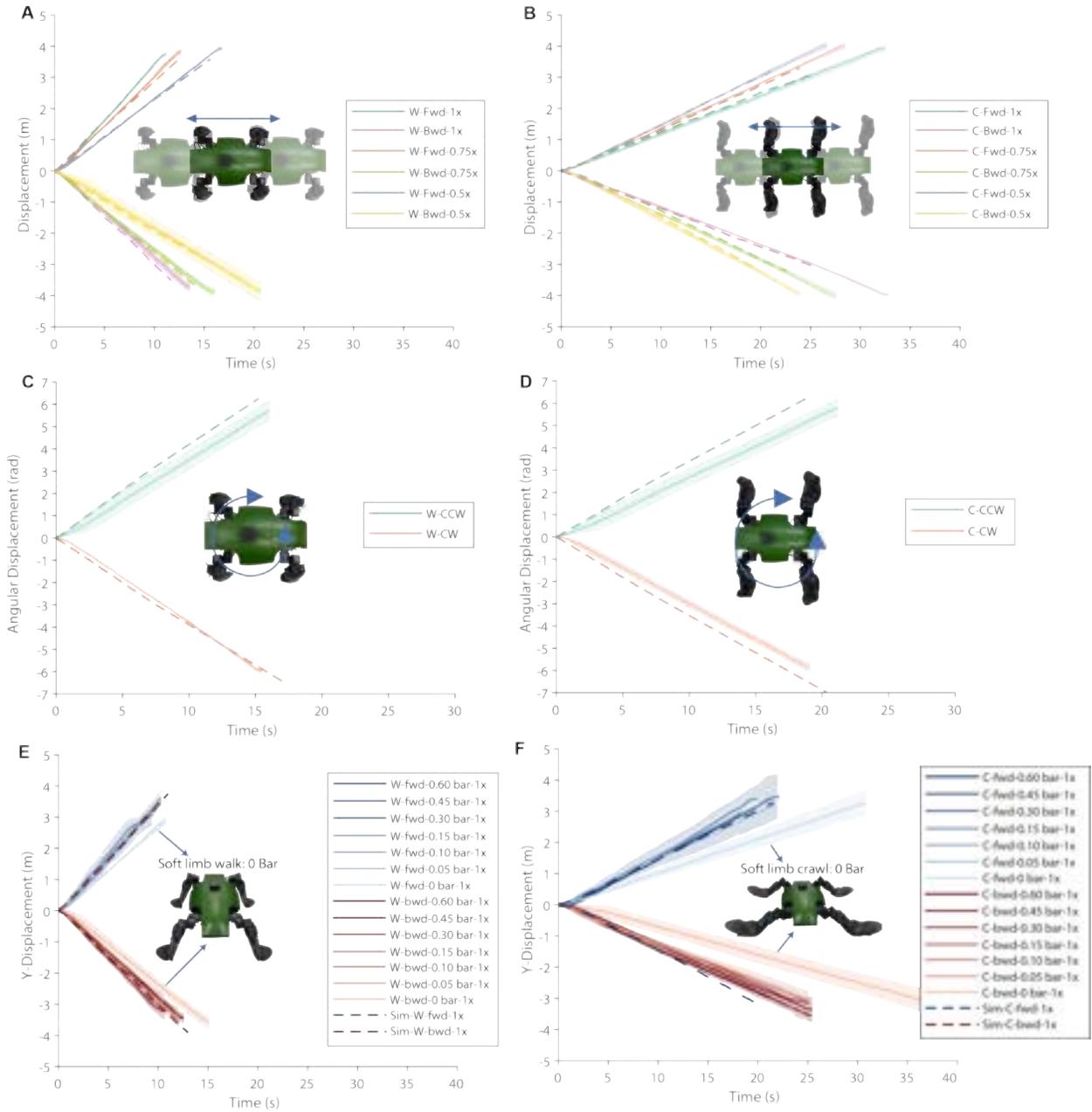

Figure 3: **Sim-to-real performance under different gaits and stiffness configurations on flat terrain.** **A,B**, Comparison of simulated (dashed) and experimental (solid) trajectories for forward and backward walking (A) and crawling (B) at commanded velocities of $V_x = \pm 0.5, \pm 0.75$, and $\pm 1$. Shaded regions indicate upper/lower bounds across five trials. **C,D**, Sim-to-real mapping of angular velocity during in-place turning for walking (C) and crawling (D); positive and negative inputs correspond to counterclockwise and clockwise rotations, respectively. **E,F**, Sim-to-real speed alignment during walking (E) and crawling (F) across varying pouch pressures.

For crawling, optimal agreement was obtained with $K_p = 800$, $K_d = 15$, and a 10 Hz update frequency. As shown in Fig. 3B, forward and backward crawling followed the same commanded velocities, producing lower slopes relative to walking—reflecting slower progression—but similarly strong correspondence between simulation and experiment: across all commanded forward and backward speeds, relative errors range from 1.00% to 5.20% (mean 3.41%) (table S2).

Angular velocity performance (Fig. 3C–D) was evaluated during pivoting, with positive values indicating counterclockwise (CCW) and negative values indicating clockwise (CW) rotation. In the walking configuration, CW rotations matched simulation closely, with a relative error of 6.3%, while CCW rotations exhibited modest under-rotation (12.2% relative error), possibly due to small geometric asymmetries in the robot. Crawling pivots showed minor offsets in both directions, with experimental angular displacement slightly suppressed relative to simulation, corresponding to relative errors of 6.69% (CW) and 16.87% (CCW) (tables S1, S2).

Lateral motion was assessed using commanded velocities of $V_y = 0.75$ (rightward) and $V_y = -0.75$ (leftward), again demonstrating strong sim-to-real agreement, with relative lateral speed errors of 3.97% and 14.03%, respectively (fig. S5 and table S1). Visual comparisons of LUART's walking and crawling gaits in simulation and experiment are provided in Movie S2.

To assess the effect of limb stiffness on sim-to-real transfer, we systematically varied the internal pressure of the inflatable pouch—the primary contributor to limb stiffness—while maintaining the jamming layers in a fully jammed state. Both walking and crawling gaits were evaluated across this range of stiffness conditions.

For the walking gait, the limb remained stable without buckling at any nonzero pouch pressure, resulting in high sim-to-real fidelity, with velocity mapping errors below 6% (Fig. 3E; table S3). The error increased to ~17% when the pouch was fully deflated ($P = 0$ psi) and lost its cylindrical structure. A complementary visualization of the data, showing sim-to-real performance as a function of limb compressive stiffness (derived from pouch pressure), is provided in fig. S6A.

For the crawling gait, performance showed a stronger dependence on pouch pressure (Fig. 3F; table S4). At all nonzero pressures, the limb maintained its cylindrical geometry and primarily exhibited bending rather than buckling. As pressure and corresponding bending stiffness decreased, bending became more pronounced, slightly reducing ELL and increasing the sim-to-real error.

Forward crawling exhibited velocity mapping errors between 5.03% and 7.82%, while backward crawling consistently produced larger discrepancies, attributed to the rearward shift of the robot's center of mass. When the pouch was fully deflated ($P = 0$ psi), the limb underwent pronounced buckling, reducing its ELL to 50% or less of its nominal value (Fig. 2B). This morphological collapse caused substantial degradation in sim-to-real correspondence, with velocity mapping errors rising to 30.29% for forward crawling and 46.79% for backward crawling. A complementary visualization of this trend, showing sim-to-real performance as a function of limb bending stiffness, is shown in fig. S6B.

## 2.4 Omnidirectional and complex trajectories

Beyond the basic locomotion modes validated in sim-to-real transfer—forward and backward walking or crawling, lateral walking, and in-place turning—LUART demonstrates maneuvering capabilities by combining multiple control inputs to produce complex trajectories.

As shown in Fig. 4A, when velocity commands are simultaneously applied in both the *x*- and *y*-directions (values of −1, −0.5, 0, 0.5, and 1), the robot achieves omnidirectional translation without inducing body rotation. For each velocity pair, three trials were performed; solid lines denote the mean trajectories, and shaded regions indicate the bounds across trials. Full trajectories are provided in Movie S3, with representative frames shown in Fig. 4B.

LUART also executes smooth, continuous curve turns by combining linear and angular velocity commands. In a teleoperated demonstration, the robot maintained a constant forward velocity of $V_x = 0.5$ while its angular velocity was continuously adjusted to trace a controlled figure-eight path (Fig. 4C and Movie S3). This result highlights the robot's ability to coordinate multiple locomotion primitives for complex motion planning.

In addition, LUART dynamically transitions between walking and crawling gaits by modifying only the body-height input in the control interface (Movie S4).

To benchmark our approach against the locomotion capabilities of other soft and soft–rigid hybrid robots reported in the literature, we summarize comparisons in Table 1. The table lists locomotion tasks demonstrated, including gait switching, bidirectional linear motion, translational and rotational turning, and omnidirectional motion, and the corresponding control strategies. Our use of indirect variables to represent material deformation in simulation enables reinforcement learning to efficiently optimize gaits, yielding successful policies across all locomotion tasks relative to prior approaches.

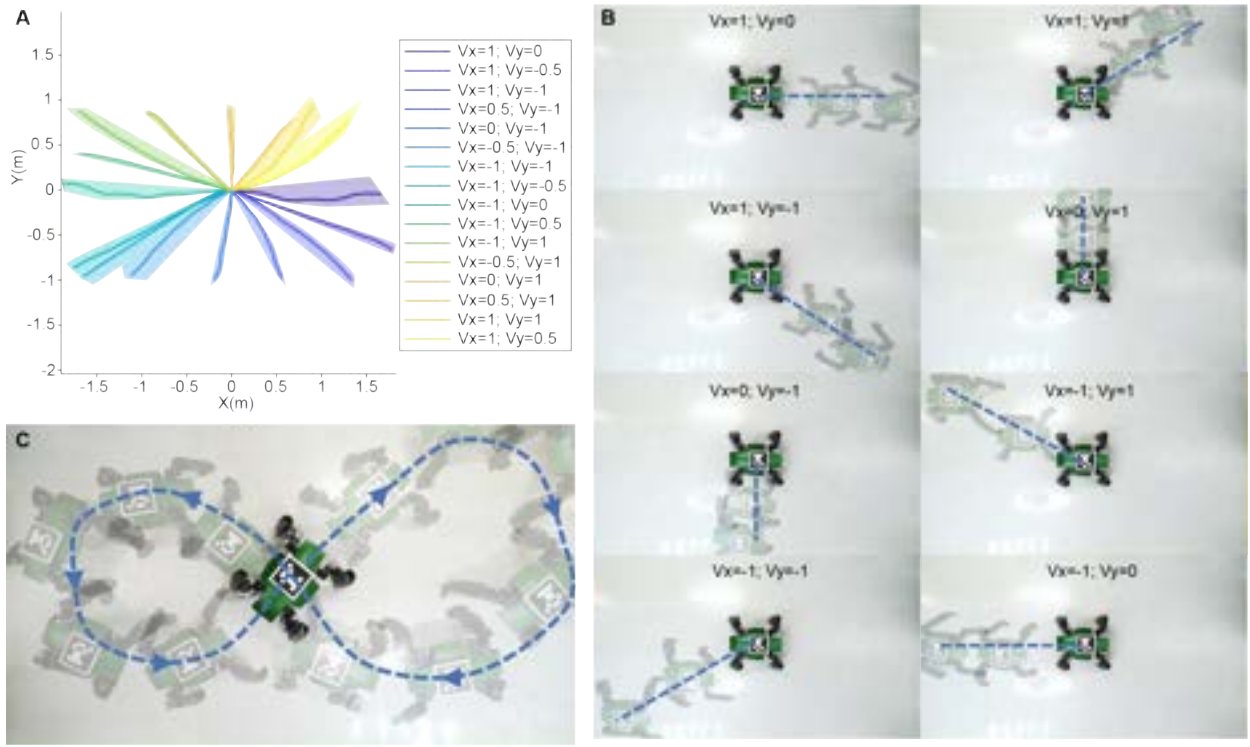

**Figure 4**: **Complex locomotion trajectories. A**, Trajectories of the robot under combined *x*- and *y*-direction velocity commands, demonstrating omnidirectional translation without body rotation. **B**, Overlaid sequential frames from corresponding trials showing representative body poses during diagonal locomotion. **C**, Figure-eight trajectory achieved by combining linear and angular velocity commands under teleoperation, illustrating smooth, coordinated curve turns.

## 2.5 Sim-to-real performance on multi-terrains

We evaluated LUART's locomotion performance across four environments: standard VCT flooring (flat, rigid terrain), artificial turf, mud, and gravel (Fig. 5A). On each terrain, we tested four locomotion modes: rigid limb walking (RLW), soft limb walking (SLW), rigid limb crawling (RLC), and soft limb crawling (SLC) (Fig. 5B). "Rigid" and "soft" denote fully inflated and deflated limb-pouch states, respectively; in both cases, the exolayers were jammed.

VCT flooring and artificial turf were treated as flat, rigid terrains, differing primarily in surface friction. The turf terrain here refers to short-pile artificial grass over a firm substrate. In contrast, the outdoor grass tested in the following section consists of long grass over rain-softened soil and behaves more like a soft, mud-like environment. Mud and gravel were categorized as soft terrains because both undergo appreciable deformation under LUART's loading.

Table 1: Locomotion tasks among soft and soft-rigid hybrid robots.

| Reference | Inspiration morphology | Multi gait | Bidirectional linear motion | Translational turns | Rotational turns | Omnidirectional linear motion | Control method |
|---|---|---|---|---|---|---|---|
| Wang et al. (11) | Snake | × | ✓ | ✓ | × | × | Model-based open-loop |
| Vaquero et al. (31) | Snake | × | ✓ | ✓ | × | × | Model-based closed-loop |
| Li et al. (32) | Snake | × | × | × | × | × | RL & sim-to-real |
| Liu et al. (33) | Snake | × | ✓ | ✓ | × | × | RL (Online learning) |
| Comoretto et al. (34) | Stick insects | × | × | ✓ | ✓ | × | Physically coupled open-loop |
| Ji et al. (35) | Legged insects | × | × | ✓ | ✓ | × | Closed-loop |
| Liang et al. (9) | Legged insects | × | × | ✓ | × | × | Open-loop |
| Shepherd et al. (8) | Worm | ✓ | × | × | × | × | Open-loop |
| Shah et al. (10) | Worm | ✓ | × | × | × | × | Open-loop |
| De Rivaz et al. (36) | Cockroach | × | × | ✓ | × | × | Open-loop |
| Das et al. (37) | Worm | ✓ | × | × | × | × | Open-loop |
| Wu et al. (38) | Caterpillar | ✓ | ✓ | × | × | × | Open-loop |
| Wu et al. (39) | Caterpillar | × | ✓ | ✓ | × | × | Open-loop |
| Tang et al. (40) | Spine | × | × | × | × | × | Open-loop |
| Ketchum et al. (41) | Quadruped | × | ✓ | × | ✓ | × | Closed-loop |
| Mazouchova et al. (42) | Turtle | ✓ | × | × | × | × | Open-loop |
| Drotman et al. (43) | Turtle | × | ✓ | × | × | × | Open-loop |
| Zhong et al. (12) | Turtle | ✓ | × | ✓ | × | × | Open-loop |
| Sun et al. (44) | Turtle | ✓ | × | ✓ | × | × | Open-loop |
| Baines et al. (3) | Turtle | ✓ | × | × | × | × | Open-loop |
| **This work** | **Turtle** | ✓ | ✓ | ✓ | ✓ | ✓ | **RL & sim-to-real** |

Figs. 5C–F show LUART's performance using rigid limbs on rigid terrains for walking and crawling gaits, including forward/backward (fwd/bwd) motion and counterclockwise/clockwise (CCW/CW) rotations. The closest agreement with simulation occurred on VCT flooring across all gaits. On artificial turf, where friction and minor surface compliance differ slightly from VCT, LUART maintained linear and repeatable trajectories, albeit with reduced speeds relative to simulation. Rigid-limb walking (RLW) speeds on turf decreased by 0.1023 m/s (29.97%, fwd) and 0.0620 m/s (20.26%, bwd), and turning rates decreased by 0.1039 rad/s (25.32%, CCW) and 0.0659 rad/s (17.50%, CW). Rigid-limb crawling (RLC) speeds decreased by 0.0238 m/s (15.32%, fwd) and 0.0237 m/s (14.32%, bwd), with turning-rate reductions of 0.1363 rad/s (41.45%, CCW) and 0.1007 rad/s (29.57%, CW).

The soft-limb walking (SLW) results on rigid terrains are shown in Figs. 5G–H. In the rigid-limb state, the tip of the limb covers the inflated cylindrical pouch and produces a rounded end-effector contact. In the soft-limb state, the uninflated pouch tip folds outward, increasing the effective ground-contact area. Soft-limb walking (SLW) performance remained comparable to rigid-limb performance: despite changes in effective limb length and limb–ground friction, the resulting gaits were linear, stable, and exhibited low variance. Deviations from simulation in fwd/bwd walking were 0.0037 m/s (1.08%, fwd) and 0.0548 m/s (17.91%, bwd) on VCT flooring, and 0.0411 m/s (12.04%, fwd) and 0.0618 m/s (20.20%, bwd) on turf. For CCW/CW turning, deviations were 0.1313 rad/s (32.00%, CCW) and 0.0492 rad/s (13.06%, CW) on VCT flooring, and 0.1441 rad/s (35.12%, CCW) and 0.0714 rad/s (18.96%, CW) on turf.

Soft-limb crawling (SLC) on rigid terrains, however, showed more substantially reduced speeds relative to simulation. Figs. 5I-J show the fwd/bwd crawling speeds decreased by 0.0471 m/s (30.31%, fwd) and 0.0782 m/s (47.25%, bwd) on VCT flooring, and by 0.0749 m/s (48.20%, fwd) and 0.1029 m/s (62.17%, bwd) on turf. Turning speeds (CCW/CW) dropped by 0.1167 rad/s (35.49%, CCW) and 0.1093 rad/s (32.09%, CW) on VCT flooring, and by 0.1286 rad/s (39.11%, CCW) and 0.1614 rad/s (47.39%, CW) on turf.

The rigid-limb walking (RLW) results on soft terrains are shown in Figs. 5K–L. On gravel, only the forward walking gait consistently reached its target, and even then, the trajectory was nonlinear, indicating intermittent sticking. The backward gait frequently stalled. On mud, both forward and backward gaits exhibited large error bars due to trials that failed midway. All turning gaits failed entirely, indicating that RLW is not well-suited for soft terrains.

The rigid-limb crawling (RLC) case on soft terrains is shown in Figs. 5M-N. The robot's lower center of mass in the crawling configuration improved overall stability. On gravel, LUART completed all crawling gaits (fwd/bwd and CCW/CW). However, on mud, we observed failures in forward and backward gaits, although turning was more consistently successful. The results suggest that RLC offers improved stability relative to RLW on soft terrains.

Soft-limb walking (SLW) on soft terrains is shown in Figs. 5O–P. SLW performance is comparable to RLW on soft substrates and frequently results in gait failure due to limb sticking. Soft-limb crawling (SLC) is consistently successful on soft terrains (Figs. 5Q-R), which we attribute to improved terrain conformity and increased contact area. This enables all crawling gaits (fwd/bwd, CCW/CW) to reach their targets with near-linear trajectories and low variance. As with other cases, SLC exhibits a modest reduction in overall locomotion speed relative to the simulated prediction. The fwd/bwd crawling speeds decreased by 0.0336 m/s (21.62%, fwd) and 0.0909 m/s (54.92%, bwd) on gravel, and by 0.0669 m/s (43.05%, fwd) and 0.1119 m/s (67.61%, bwd) on mud. Turning speeds (CCW/CW) dropped by 0.0671 rad/s (20.41%, CCW) and 0.0602 rad/s (17.67%, CW) on gravel, and by 0.0854 rad/s (25.97%, CCW) and 0.1089 rad/s (31.97%, CW) on mud.

fig. S7 summarizes the COT across terrains for all four limb–gait–terrain combinations. For in-place turning, angular velocity was used to compute COT analogously to linear velocity. On rigid terrains, LUART's walking gaits (RLW and SLW) achieved both higher speeds and lower COT than crawling. On soft terrains, crawling gaits (RLC and SLC) proved more effective. Compared with RLC, SLC produced more stable locomotion on soft terrains, resulting in a lower average COT on both mud and gravel. From both sim-to-real performance and energetic standpoints, SLC emerges as the optimal locomotion strategy for soft, deformable terrains. Compared to our previous amphibious robotic turtle, which relied on open-loop and hand-tuned gaits (*3*), LUART further achieves an order-of-magnitude reduction in COT, highlighting the efficiency gains enabled by closed-loop and RL-defined gaits.

Lateral motions were analyzed across all limb–gait–terrain configurations and are shown in

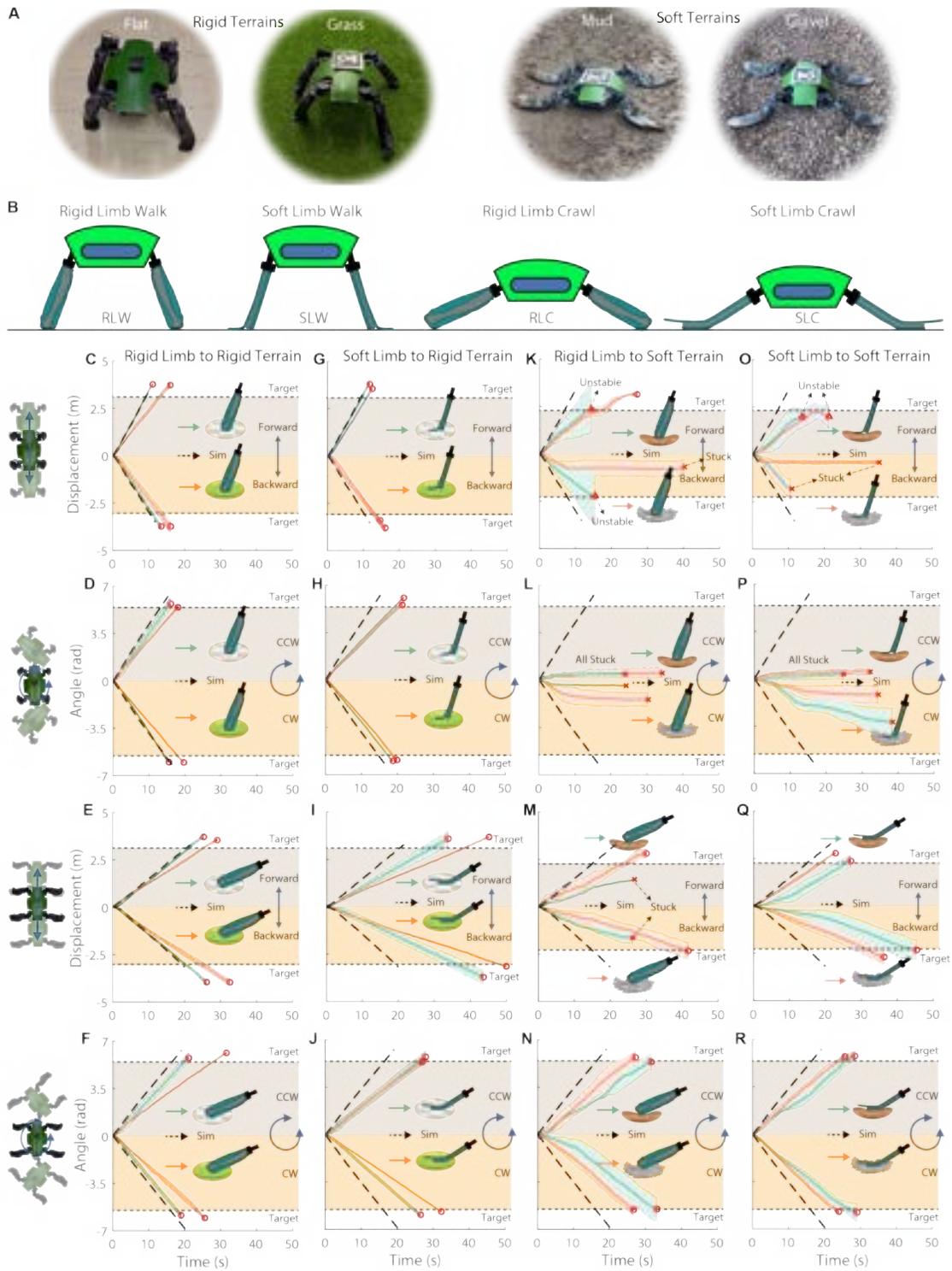

**Figure 5**: **Sim-to-real locomotion performance of LUART across terrains, limb states, and gaits. A**, Four evaluated terrains: flat VCT flooring, artificial turf, mud, and gravel, grouped into rigid (flat, turf) and soft (mud, gravel) classes. **B**, Four locomotion modes: rigid-limb walking (RLW), soft-limb walking (SLW), rigid-limb crawling (RLC), and soft-limb crawling (SLC). **C–F**, Rigid limbs on rigid terrains. **G–J**, Soft limbs on rigid terrains. **K–N**, Rigid limbs on soft terrains. **O–R**, Soft limbs on soft terrains. Within each block: first column, forward/backward walking; second, in-place turning during walking; third, forward/backward crawling; fourth, in-place turning during crawling. See Movie S5 for real-time demonstrations.

figs. S8A–B. Lateral movement is only feasible in the walking gait and therefore could not be executed on the soft terrains. On the rigid terrains, the robot achieved stable lateral translation. The corresponding COT values are presented in figs. S8C–D.

The performance of all limb configurations on both soft and rigid terrains is shown in Movie S5.

## 2.6 Field demonstrations

We demonstrate LUART's maneuverability in a real-world outdoor environment. We constructed a multi-terrain course featuring flat concrete, mud, grass, gravel, and a bench obstacle to test teleoperated navigation (Movie S6). Unlike the controlled indoor setup that relied on ArUco tracking, outdoor position was obtained via GPS and real-time power via an INA sensor, enabling online COT estimation (hardware details are provided in the Methods).

We first ran a trial involving both limb-mode and gait transitions. LUART began with rigid-limb walking (RLW) on concrete. When approaching the mud and grass segments, the limbs were deflated, and upon entering the soft terrain the robot lowered its body height to transition into soft-limb crawling (SLC). It then continued onto gravel, crawled beneath a bench, traversed an additional mud–grass segment, reinflated the limbs, and switched back to walking to complete the course. Ghosted frames from this run are shown in Figs. 6A–C. Terrain-specific COT values (Fig. 6D) show that RLW on flat concrete achieved an average COT of ~2, while SLC maintained an average COT of ~4 across mud, grass, and gravel.

To contextualize the benefits of adaptive limb and gait switching, we conducted two control trials representing two limiting cases: (i) pure RLW, which is expected to perform well only on firm, structured terrain, and (ii) pure rigid-limb crawling (RLC), which lowers the center of mass and improves stability but is expected to decrease locomotion efficiency across all terrains. As shown in Fig. 6B and Movie S6, RLW traversed concrete and grass but became stuck on gravel. Although RLW exhibited lower COT than SLC on mud and grass, the complete stall on gravel caused a large spike in COT. In contrast, the pure RLC trial (Fig. 6C) completed the entire course, but its COT was markedly increased relative to RLW and SLC; on flat and gravel terrain, the COT corresponding to RLC was more than twice that of RLW and SLC, respectively (Fig. 6D).

Fig. 6D summarizes terrain-specific COT for all three trials, with full online COT traces in

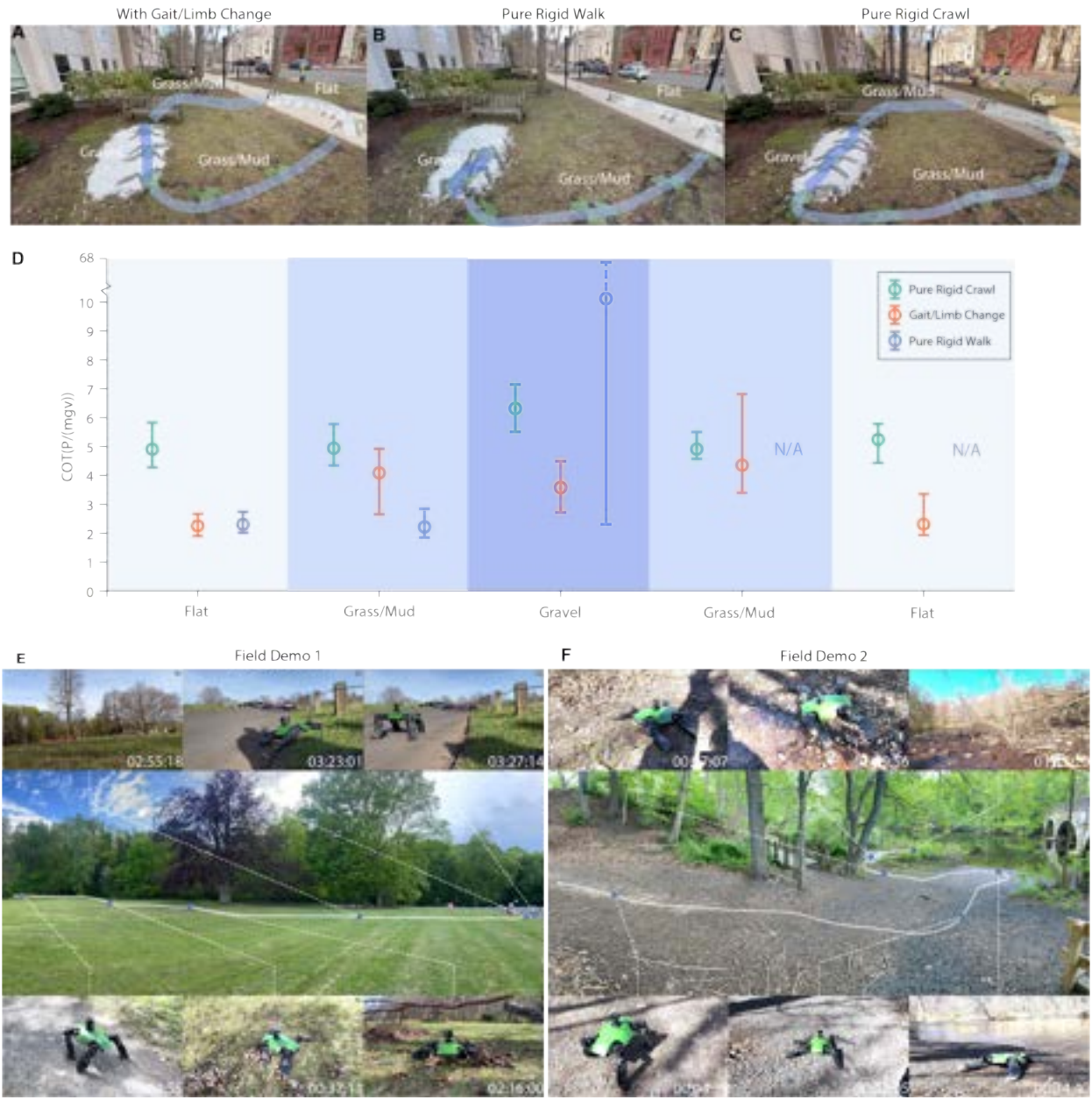

**Figure 6**: **Field demonstrations of multi-terrain locomotion. A–C**, Ghosted frames from the Yale campus multi-terrain trial showing LUART traversing flat concrete, mud and grass, gravel, and returning to flat terrain. Three conditions were tested: limb-mode and gait switching (**A**), pure rigid-limb walking (**B**), and pure rigid-limb crawling (**C**). **D**, Online COT measurements for each terrain segment, with error bars indicating variability across runs. Limb-mode and gait adaptation achieved the lowest overall COT while maintaining mobility. **F**, Field test at East Rock Park: LUART transitioned from rigid-limb walking on a pedestrian trail to soft-limb crawling over long, wet grass, descended a slope, crossed a natural branch obstacle, and climbed a curb before returning to walking. **G**, Field test along an estuary: LUART performed soft-limb crawling over gravel, mud, leaves, and shallow water, including slope climbing and backward reorientation to navigate confined obstacles.

fig. S9. Limb inflation and deflation introduced small, momentary COT increases, but these were negligible compared to the substantial energy and robustness benefits gained through limb-mode and gait adaptations.

We also conducted long-range field tests in more complex, unstructured outdoor environments. The first test took place at East Rock Park in New Haven, CT. LUART began with rigid-limb walking on a compact pedestrian path, then transitioned onto a large grassy area with long grass and rain-softened soil. Upon entering this terrain, it switched to soft-limb crawling. We teleoperated LUART down a mild grassy slope toward a parking lot, where it successfully navigated over a naturally formed arch-shaped obstacle created by fallen branches. After reaching the lot, it climbed a curb and returned to rigid-limb walking to continue its route (Fig. 6E and Movie S7).

The second field test was conducted along an estuary. LUART started on hard-packed terrain using rigid limbs and switched to soft-limb crawling upon reaching the gravel-covered riverbank. It then traversed the shoreline and entered a shallow streambed composed of mud, leaves, and water. In this segment, LUART climbed small slopes and, when faced with narrow or misaligned obstacles, used its backward gait to reorient and navigate through confined spaces (Fig. 6F and Movie S8).

## 3  Discussion

We have demonstrated that indirect modeling of soft-body deformation within a rigid-body simulator enables reinforcement learning to discover efficacious and energy-efficient locomotion policies for partially soft (and potentially all-soft) robots. We applied this approach to a quadruped robot with soft, morphing limbs. By randomizing ELLs and centers of mass in simulation, the derived locomotion policies were found to be inherently robust to variability in limb compliance and geometric uncertainty, allowing high-fidelity sim-to-real transfer (with velocity mapping errors below 6%) across a spectrum of locomotion tasks.

Field experiments further show the practical benefits of the surrogate compliance modeling approach. LUART successfully traversed multi-terrain courses, including obstacles and unstructured environments, demonstrating that the learned policies generalize beyond controlled laboratory conditions. Notably, the robot maintained a low cost of transport during dynamic maneuvers, including crawling under a bench, ascending curbs, and reorienting in confined spaces. Overall, the RL-discovered gaits achieved COTs an order of magnitude lower than previous hand-designed gaits, suggesting that RL-discovered gaits and limb-morphology adaptations together support efficient multi-environment locomotion.

While our indirect soft-body modeling approach offers substantial advantages, several limitations and opportunities remain. First, sim-to-real fidelity decreases on highly deformable substrates or when limb compliance exceeds the range captured in training, indicating that broader or online-adaptive randomization could further improve robustness. Second, the current study focuses on terrestrial locomotion; future work could extend this methodology to multi-modal locomotion in aquatic environments. Finally, integrating advanced proprioceptive and exteroceptive sensing, such as dynamic slip detection or nuanced terrain classification algorithms, could enhance real-time adaptability and extend LUART's operational envelope.

By capturing the essential mechanical effects of compliance, rather than reproducing full soft-body physics, our surrogate compliance modeling approach makes behavioral optimization of soft robots tractable in rigid-body simulation. This strategy generalizes across morphogenetic and soft robotic platforms, providing an accessible, simulation-ready path to robust gait optimization and improved field readiness.

## Acknowledgments

We thank F. Fish for support and for providing video materials. We acknowledge the Philadelphia Zoo, the Leatherback Trust, and the Goldring-Gund Marine Biology Station, Playa Grande, Santa Cruz, Costa Rica, for their assistance. Video footage of the Olive Ridley sea turtle (*Lepidochelys olivacea*) was recorded at the National Wildlife Refuge Ostional under permit number ACT-OR-DR-089-2025, provided by the Tempisque Conservation Area (ACT). We also thank V. Valverde, D. Kramer, S. Sebo, A. Tilley, and M. Williamson for their assistance.

**Funding:** This project was sponsored by the Office of Naval Research under award N00014-24-1-2162. Any opinions, findings, and conclusions or recommendations expressed in this material are those of the authors and do not necessarily reflect the views of the Office of Naval Research.

L.A.R. was partially supported by an NSF Graduate Research Fellowship (DGE-1752134).

**Author contributions:** J.W. and M.J. conceived the idea, designed the research, performed the experiments, and wrote and revised the manuscript. L.A.R., B.Y., M.Z., E.F., and W.Y. assisted with experimental execution. R.K.B. conceived the idea, revised the manuscript, and acquired funding.

**Competing interests:** The authors declare that they have no competing interests.

**Data and materials availability:** The code and data for this study are archived at `https://github.com/the-faboratory/LUART.git`.

# Supplementary materials

Materials and Methods

Supplementary Text S1

Figures S1 to S15

Tables S1 to S6

Captions for Movies S1 to S9

# Materials and Methods

## 3.1 Robot hardware

Each LUART limb is actuated by three Dynamixel motors controlling pitch, roll, and yaw (fig. S1). The primary rotational joint uses a Dynamixel XH540-W150-R; the remaining two joints use XH540-W270-R motors for increased torque.

The main body houses a Jetson Orin Nano edge computer, which performs RL inference and manages communication with sensors and the Arduino microcontroller. Power is supplied by a 14.8 V battery. IMU data are obtained from an Intel RealSense D435 camera connected to the Jetson, and encoder/control signals are transmitted over RS485 using a U2D2 converter.

A dedicated Arduino-based control unit manages pneumatic actuation for both positive pouch pressure and negative jamming pressure. It also handles communication with the GPS module, depth sensor, and power sensor before forwarding aggregated data to the Jetson.

The control architecture is shown in fig. S10. An Arduino Nano drives two 12 V motor drivers to actuate Dynoflow 6000 Series diaphragm pumps capable of both inflation and vacuum generation. Airflow is controlled by 8 mm Parker X-Valve solenoid valves, switched via transistor arrays. A Honeywell ABP Series analog pressure sensor provides real-time pouch pressure for closed-loop control.

The GPS, depth sensor, and power sensor communicate with the Arduino over $I^2C$, each with a unique address. The SparkFun NEO-M9N GPS module is paired with a signal amplifier; depth is measured using the SparkFun VL53L4CD (1 mm resolution); and current/voltage is monitored using the INA219.

The PCB layout and implementation are shown in fig. S11. fig. S11A–B show the front and back layouts. fig. S11C highlights the assembled front side, containing the power sensor, pump drivers, transistor arrays, and both positive and negative pressure sensors (monitoring pouch pressure and jamming vacuum). A 4-pin connector interfaces with the depth sensor mounted on the robot underside. A dual-purpose pump inflates the pouch and evacuates the jamming layer. fig. S11D shows the back side, including the GPS module, solenoid valve array, and Arduino Nano.

## 3.2 Stiffness-tunable and morphing limbs

The limb consists of three components (fig. S12). Parts A and B form a sandwich structure made of two 1/8-inch latex sheets encapsulating ten layers of PET kirigami-patterned film that form the jamming exolayers (*29, 45*). A silicone wall between the latex sheets creates the sealed jamming cavity, bonded using Sil-Poxy. Openings for silicone tubing are sealed with Sil-Poxy to ensure airtightness.

Between Parts A and B lies an inflatable pouch (fig. S13), fabricated from PVC vinyl-coated polyester. A 240 mm × 240 mm sheet is cut, coated along the edges with HH-66 vinyl cement, folded to form a cavity, and Z-folded into a 40 mm × 240 mm pouch. A vinyl-coated polyester hood is bonded to one end as a fixed anchor. An air inlet is added and sealed with super glue and hot glue. The pouch is leak-tested by water immersion and patched if necessary.

After preparing the components, a steel cable is attached to the pouch's free end and routed around an extended segment of Part B. When the pouch inflates and shortens, the cable pulls the overhanging section downward, bending the limb into a cylindrical, rigid-like shape. The cable is secured using two latex strips.

Parts A and B are then sewn together. To allow free sliding of the pouch, it is wrapped with a thin PVC film. A circular housing between Parts A and B anchors the hooded end of the pouch via a carbon fiber rod. During inflation, only the cable-linked free end contracts. Two clamps secure the latex layers, and the assembly is mounted to the motor unit.

A demonstration of switching between leg and flipper modes via pouch pressure control is provided in Movie S9.

## 3.3 Reinforcement learning model training

We trained a locomotion policy for our quadruped robot using Proximal Policy Optimization (PPO), implemented with the `rsl_rl` library inside the Genesis GPU-accelerated simulation environment (*46*). To enable efficient data collection and rapid policy optimization, training was performed in a massively parallel fashion, deploying up to 10,000 simulated robot instances running synchronously.

Our approach employed a teacher-student training framework, in which the teacher policy was granted privileged information, such as full state and latent environment variables, to accelerate learning, while the deployable student policy relied solely on realistic onboard sensor data. Both the actor and critic networks consisted of three fully connected layers with hidden sizes of [512, 256, 128] and ELU activation functions, sharing a common backbone with separate output heads for the student and teacher. The Gaussian policy's output standard deviation was initialized to 1.0 and annealed throughout training to stabilize exploration.

The simulated robotic turtle was modeled using a modified URDF, with eight actuated degrees of freedom and a floating base. As shown in fig. S14, the base connects to four limbs, each comprising three serially connected cuboid segments that represent the three actuated joints responsible for pitch, roll, and yaw motion. These cuboids serve as both visual and collision elements, simplifying the representation of the limb kinematics. Each limb's collision geometry was approximated as a regular cylinder capped with a fixed hemispherical shell at the distal end to facilitate stable and smooth ground contact during locomotion. Simulation dynamics were integrated at 1 kHz, and control signals were applied at 25 Hz (a control decimation factor of 40) to reflect real-world system latency. Each training episode lasted 20 seconds, during which the command velocities (forward, lateral, and yaw angular velocities, as well as target body height) were randomized and held constant for 4 second intervals to promote both stability and adaptability. Observations for the policy network were constructed by concatenating angular velocity, gravity vector (in the base frame), joint positions and velocities, height sensor data, and a window of past actions, with additional noise injected into all proprioceptive and height sensing channels to further promote robustness.

To enhance policy generalization and facilitate sim-to-real transfer, we implemented extensive domain randomization. At the start of each episode, limb lengths were randomly scaled between 0.6 and 1.1 times the nominal value, generating a diverse set of 10,000 robot morphologies. The center of mass could be shifted up to ±25% of the limb length, and the base mass was perturbed within a range of −1 to +2 kg. Ground friction coefficients were uniformly sampled from 0.2 to 1.5 to capture a wide variety of surface conditions. We also randomized actuator and controller parameters, including motor output offsets, PD controller gains ($K_p$ and $K_d$), joint stiffness, damping, and initial joint positions, ensuring the learned policy would be robust to hardware and environmental variability.

The reward function comprised positive terms for tracking the commanded linear velocity and yaw rate, along with proper foot placement and air time. Penalties were applied to deviations from the target base height and orientation (roll and pitch) to discourage instability. Additional penalties on total torque and rapid changes in action output (the `torques` and `action_rate` terms) limited energy use and promoted smooth, efficient motion. Negative rewards were also assigned for collisions, undesired vertical velocity, excessive angular velocities, and premature foot contact. We further extended the reward function with custom terms on base pitch, vertical lift-off velocity, joint symmetry, and foot positioning. Training reward definitions are provided in table S5.

Optimization was performed using PPO with a clipped surrogate loss (`clip_param`=0.2), generalized advantage estimation ($\lambda = 0.95$, $\gamma = 0.99$), and an adaptive KL penalty (target = 0.01) for stability. Each training iteration included 24 steps per environment and 5 epochs of minibatch updates, with gradients clipped to a maximum norm of 1.0. The value loss coefficient was set to 1.0, and an entropy coefficient of 0.01 encouraged sufficient exploration. The learning rate was set to $1 \times 10^{-3}$ and adaptively scheduled during training. All experiments were conducted with a fixed random seed and model checkpoints were saved every 100 iterations. Training progress was periodically evaluated using deterministic rollouts in randomized physical environments. Training hyperparameters are shown in table S6.

The training results demonstrated rapid and stable convergence of all key reward components (fig. S15). Both the linear velocity reward and angular velocity reward increased sharply and plateaued within the first 200 training iterations, indicating that the agent efficiently learned to track commanded translational and rotational velocities. The total reward followed a similar trend, exhibiting a steep rise during the initial training phase and stabilizing thereafter. In the absence of URDF randomization, a complete run of 1000 iterations required approximately 600 seconds on an NVIDIA RTX 4090 GPU.

# References and Notes

1. D. S. Shah, M. C. Yuen, L. G. Tilton, E. J. Yang, R. Kramer-Bottiglio, Morphing robots using robotic skins that sculpt clay. *IEEE Robotics and Automation Letters* **4** (2), 2204–2211 (2019).

2. R. Baines, F. Fish, J. Bongard, R. Kramer-Bottiglio, Robots that evolve on demand. *Nature Reviews Materials* **9** (11), 822–835 (2024).

3. R. Baines, *et al.*, Multi-environment robotic transitions through adaptive morphogenesis. *Nature* **610** (7931), 283–289 (2022).

4. T. Kim, S. Lee, S. Chang, S. Hwang, Y.-L. Park, Environmental adaptability of legged robots with cutaneous inflation and sensation. *Advanced Intelligent Systems* **5** (11), 2300172 (2023).

5. D. Melancon, B. Gorissen, C. J. García-Mora, C. Hoberman, K. Bertoldi, Multistable inflatable origami structures at the metre scale. *Nature* **592** (7855), 545–550 (2021).

6. H. Kabutz, A. Hedrick, W. P. McDonnell, K. Jayaram, mCLARI: a shape-morphing insect-scale robot capable of omnidirectional terrain-adaptive locomotion in laterally confined spaces, in *2023 IEEE/RSJ International Conference on Intelligent Robots and Systems (IROS)* (IEEE) (2023), pp. 8371–8376.

7. T. F. Nygaard, C. P. Martin, J. Torresen, K. Glette, D. Howard, Real-world embodied AI through a morphologically adaptive quadruped robot. *Nature Machine Intelligence* **3** (5), 410–419 (2021).

8. R. F. Shepherd, *et al.*, Multigait soft robot. *Proceedings of the National Academy of Sciences* **108** (51), 20400–20403 (2011), publisher: Proceedings of the National Academy of Sciences, doi:10.1073/pnas.1116564108, https://pnas.org/doi/full/10.1073/pnas.1116564108.

9. J. Liang, *et al.*, Electrostatic footpads enable agile insect-scale soft robots with trajectory control. *Science Robotics* **6** (55) (2021), publisher: American Association for the Advancement of Science (AAAS), doi:10.1126/scirobotics.abe7906, https://www.science.org/doi/10.1126/scirobotics.abe7906.

10. D. S. Shah, *et al.*, A soft robot that adapts to environments through shape change. *Nature Machine Intelligence* **3** (1), 51–59 (2020), publisher: Springer Science and Business Media LLC, doi:10.1038/s42256-020-00263-1, https://www.nature.com/articles/s42256-020-00263-1.

11. T. Wang, *et al.*, Mechanical intelligence simplifies control in terrestrial limbless locomotion. *Science Robotics* **8** (85) (2023), publisher: American Association for the Advancement of Science (AAAS), doi:10.1126/scirobotics.adi2243, https://www.science.org/doi/10.1126/scirobotics.adi2243.

12. W. Zhong, Y. Wu, L. Li, J. Shao, X. Gu, A tortoise-inspired quadrupedal pneumatic soft robot that adapts to environments through shape change. *Bioinspiration & Biomimetics* **20** (3), 036002 (2025), publisher: IOP Publishing, doi:10.1088/1748-3190/adbc5d, https://iopscience.iop.org/article/10.1088/1748-3190/adbc5d.

13. J. Hwangbo, *et al.*, Learning agile and dynamic motor skills for legged robots. *Science Robotics* **4** (26), eaau5872 (2019).

14. T. Miki, *et al.*, Learning robust perceptive locomotion for quadrupedal robots in the wild. *Science robotics* **7** (62), eabk2822 (2022).

15. Y. Ma, A. Cramariuc, F. Farshidian, M. Hutter, Learning coordinated badminton skills for legged manipulators. *Science Robotics* **10** (102), eadu3922 (2025).

16. H. Kim, *et al.*, High-speed control and navigation for quadrupedal robots on complex and discrete terrain. *Science Robotics* **10** (102), eads6192 (2025).

17. A. Liu, *et al.*, An Intelligent Bionic Amphibious Turtle Robot With Visual-Tactile Fusion for Dynamic Terrain Adaptation. *IEEE Transactions on Robotics* **41**, 6345–6363 (2025).

18. X. Liu, C. D. Onal, J. Fu, Reinforcement learning of CPG-regulated locomotion controller for a soft snake robot. *IEEE Transactions on Robotics* **39** (5), 3382–3401 (2023).

19. G. Li, J. Shintake, M. Hayashibe, Deep reinforcement learning framework for underwater locomotion of soft robot, in *2021 IEEE international conference on robotics and automation (ICRA)* (IEEE) (2021), pp. 12033–12039.

20. C. M. Hubicki, J. J. Aguilar, D. I. Goldman, A. D. Ames, Tractable terrain-aware motion planning on granular media: An impulsive jumping study, in *2016 IEEE/RSJ International Conference on Intelligent Robots and Systems (IROS)* (IEEE) (2016), pp. 3887–3892.

21. D. J. Lynch, K. M. Lynch, P. B. Umbanhowar, The soft-landing problem: Minimizing energy loss by a legged robot impacting yielding terrain. *IEEE Robotics and Automation Letters* **5** (2), 3658–3665 (2020).

22. Y. Su, *et al.*, A unified foot–terrain interaction model for legged robots contacting with diverse terrains. *IEEE/ASME Transactions on Mechatronics* **29** (4), 2661–2672 (2023).

23. J. Lee, J. Hwangbo, L. Wellhausen, V. Koltun, M. Hutter, Learning quadrupedal locomotion over challenging terrain. *Science robotics* **5** (47), eabc5986 (2020).

24. S. Cuomo, *et al.*, Scientific machine learning through physics–informed neural networks: Where we are and what's next. *Journal of Scientific Computing* **92** (3), 88 (2022).

25. S. Choi, *et al.*, Learning quadrupedal locomotion on deformable terrain. *Science Robotics* **8** (74), eade2256 (2023).

26. C. Yao, G. Shi, Y. Ge, Z. Zhu, Z. Jia, Predict the physics-informed terrain properties over deformable soils using sensorized foot for quadruped robots, in *2023 International Conference on Advanced Robotics and Mechatronics (ICARM)* (IEEE) (2023), pp. 330–335.

27. H. Li, *et al.*, Physics-informed Neural Network Predictive Control for Quadruped Locomotion. *arXiv preprint arXiv:2503.06995* (2025).

28. D. Drotman, S. Jadhav, M. Karimi, P. de Zonia, M. T. Tolley, 3D printed soft actuators for a legged robot capable of navigating unstructured terrain, in *2017 IEEE international conference on robotics and automation (ICRA)* (IEEE) (2017), pp. 5532–5538.

29. J. Sun, *et al.*, Performance Enhancement of a Morphing Limb for an Amphibious Robotic Turtle, in *2024 IEEE 7th International Conference on Soft Robotics (RoboSoft)* (IEEE) (2024), pp. 374–379.

30. L. A. Ramirez, R. Baines, B. Yang, R. Kramer-Bottiglio, Decreasing the Cost of Morphing in Adaptive Morphogenetic Robots. *Advanced Intelligent Systems* p. 2401055 (2025).

31. T. S. Vaquero, *et al.*, EELS: Autonomous snake-like robot with task and motion planning capabilities for ice world exploration. *Science Robotics* **9** (88) (2024), publisher: American Association for the Advancement of Science (AAAS), doi:10.1126/scirobotics.adh8332, https://www.science.org/doi/10.1126/scirobotics.adh8332.

32. G. Li, J. Shintake, M. Hayashibe, Deep Reinforcement Learning Framework for Underwater Locomotion of Soft Robot, in *2021 IEEE International Conference on Robotics and Automation (ICRA)* (IEEE, Xi'an, China) (2021), pp. 12033–12039, doi:10.1109/icra48506.2021.9561145, https://ieeexplore.ieee.org/document/9561145/.

33. X. Liu, C. D. Onal, J. Fu, Reinforcement Learning of CPG-Regulated Locomotion Controller for a Soft Snake Robot. *IEEE Transactions on Robotics* **39** (5), 3382–3401 (2023), publisher: Institute of Electrical and Electronics Engineers (IEEE), doi:10.1109/tro.2023.3286046, https://ieeexplore.ieee.org/document/10175020/.

34. A. Comoretto, H. A. H. Schomaker, J. T. B. Overvelde, Physical synchronization of soft self-oscillating limbs for fast and autonomous locomotion .

35. X. Ji, *et al.*, An autonomous untethered fast soft robotic insect driven by low-voltage dielectric elastomer actuators. *Science Robotics* **4** (37) (2019), publisher: American Association for the Advancement of Science (AAAS), doi:10.1126/scirobotics.aaz6451, https://www.science.org/doi/10.1126/scirobotics.aaz6451.

36. S. D. De Rivaz, *et al.*, Inverted and vertical climbing of a quadrupedal microrobot using electroadhesion. *Science Robotics* **3** (25) (2018), publisher: American Association for the Advancement of Science (AAAS), doi:10.1126/scirobotics.aau3038, https://www.science.org/doi/10.1126/scirobotics.aau3038.

37. R. Das, S. P. M. Babu, F. Visentin, S. Palagi, B. Mazzolai, An earthworm-like modular soft robot for locomotion in multi-terrain environments. *Scientific Reports* **13** (1), 1571 (2023).

38. S. Wu, Y. Hong, Y. Zhao, J. Yin, Y. Zhu, Caterpillar-inspired soft crawling robot with distributed programmable thermal actuation. *Science Advances* **9** (12), eadf8014 (2023).
S8

39. S. Wu, T. Zhao, Y. Zhu, G. H. Paulino, Modular multi-degree-of-freedom soft origami robots with reprogrammable electrothermal actuation. *Proceedings of the National Academy of Sciences* **121** (20) (2024), publisher: Proceedings of the National Academy of Sciences, doi: 10.1073/pnas.2322625121, https://pnas.org/doi/10.1073/pnas.2322625121.

40. Y. Tang, *et al.*, Leveraging elastic instabilities for amplified performance: Spine-inspired high-speed and high-force soft robots. *SCIENCE ADVANCES* (2020).

41. J. Ketchum, *et al.*, Automated Gait Generation for Walking, Soft Robotic Quadrupeds, in *2023 IEEE/RSJ International Conference on Intelligent Robots and Systems (IROS)* (IEEE, Detroit, MI, USA) (2023), pp. 10245–10251, doi:10.1109/iros55552.2023.10342059, https://ieeexplore.ieee.org/document/10342059/.

42. N. Mazouchova, P. B. Umbanhowar, D. I. Goldman, Flipper-driven terrestrial locomotion of a sea turtle-inspired robot. *Bioinspiration & Biomimetics* **8** (2), 026007 (2013), publisher: IOP Publishing, doi:10.1088/1748-3182/8/2/026007, https://iopscience.iop.org/article/10.1088/1748-3182/8/2/026007.

43. D. Drotman, S. Jadhav, D. Sharp, C. Chan, M. T. Tolley, Electronics-free pneumatic circuits for controlling soft-legged robots. *Science Robotics* **6** (51) (2021), publisher: American Association for the Advancement of Science (AAAS), doi:10.1126/scirobotics.aay2627, https://www.science.org/doi/10.1126/scirobotics.aay2627.

44. L. Sun, J. Wan, T. Du, Fully 3D-printed tortoise-like soft mobile robot with muti-scenario adaptability. *Bioinspiration & Biomimetics* **18** (6), 066011 (2023), publisher: IOP Publishing, doi:10.1088/1748-3190/acfd76, https://iopscience.iop.org/article/10.1088/1748-3190/acfd76.

45. R. Baines, B. Yang, L. A. Ramirez, R. Kramer-Bottiglio, Kirigami layer jamming. *Extreme Mechanics Letters* **64**, 102084 (2023).

46. G. Authors, Genesis: A Universal and Generative Physics Engine for Robotics and Beyond (2024), https://github.com/Genesis-Embodied-AI/Genesis.